\documentclass{article} 
\usepackage{iclr2016_conference,times}
\usepackage{hyperref}
\usepackage{url}
\usepackage{amsmath}
\usepackage{comment}
\usepackage{tabularx}
\usepackage{capt-of}
\usepackage{graphicx}
\usepackage{multirow}
\usepackage{subfig}
\usepackage{booktabs}
\usepackage{authblk, etoolbox}

\usepackage{blindtext}


\title{Tracking the World State with \\ Recurrent Entity Networks}

\newcommand{\modelname}{Recurrent Entity Network}
\newcommand{\modelabbrev}{EntNet}

\author[1,2]{Mikael Henaff}
\author[1]{Jason Weston}
\author[1]{Arthur Szlam}
\author[1]{Antoine Bordes}
\author[1,2]{Yann LeCun}

\affil[1]{Facebook AI Research}
\affil[2]{Courant Institute, New York University}
\affil[ ]{\texttt{\{mbh305\}@nyu.edu}, \texttt{\{jase,aszlam,abordes,yann\}@fb.com}}

\iclrfinalcopy 

\begin{document}

\makeatletter
\patchcmd{\@maketitle}{center}{flushleft}{}{}
\patchcmd{\@maketitle}{center}{flushleft}{}{}

\def\maketitle{{%
  \renewenvironment{tabular}[2][]
    {\begin{flushleft}}
    {\end{flushleft}}
  \AB@maketitle}}
\makeatother

\renewcommand\Authands{ and }
\renewcommand\Authfont{\normalfont\normalsize}
\renewcommand\Affilfont{\normalfont\normalsize}

\renewcommand*{\Authfont}{\bfseries}

\maketitle

\begin{abstract}
We introduce a new model, the \modelname~(\modelabbrev). 
It is equipped with a dynamic long-term memory which allows it to maintain and update a representation of the state of the world as it receives new data.
For language understanding tasks, it can reason on-the-fly as it reads text, not
just when it is required to answer a question or respond as is the case for a Memory Network \citep{MemN2N}.
Like a Neural Turing Machine or Differentiable Neural
Computer \citep{NTM, graves2016hybrid} it maintains a fixed size memory and
can learn to perform location and content-based read and write
operations.  However, unlike those models it has a simple parallel 
architecture in which several memory locations can be updated
simultaneously. The \modelabbrev~sets a new state-of-the-art on the bAbI
tasks, and is the first method to solve all the tasks in the 10k
training examples setting.  
We also demonstrate that it can solve a reasoning task which requires a large number of supporting facts, which other methods are not able to solve, and can generalize past its training horizon.
It can also be practically used
on large scale datasets such as 
Children's Book Test, where it obtains competitive performance, reading the story in a single pass.
\end{abstract}

\section{Introduction}


The essence of intelligence is the ability to predict.
An intelligent agent must be able to predict unobserved facts about their environment from limited percepts (visual, auditory, textual, or otherwise), combined with their knowledge of the past. 
In order to reason and plan, they must be able to predict how an observed event or action will affect the state of the world.
Arguably, the ability to maintain an estimate of the current state of the world, combined with a forward model of how the world evolves, is a key feature of intelligent agents.
A natural way for an agent to represent the world is to maintain a set of high-level concepts or entities together with their properties, which are updated as new information is received.
For example, if a percept is the textual description of an event, such
as ``John walks out of the kitchen'', the agent should learn to update
its estimate of John's location, as well as the list (and number) of
people present in each room.  If John was carrying a bag, the location
of the bag and the list of objects in the kitchen must also be
updated. When we read a story, each sentence we read or hear causes us
to update our internal representation of the current state of the
world within the story. The flow of the story is captured by the
evolution of this state of the world.

At any given time, an agent typically receives limited information about the state of the world, and should therefore be able to infer new information through partial observation. In this paper, we investigate this problem through a simple story understanding scenario, in which the agent is given a
sequence of textual statements and events, and then given another series of
statements about the final state of the world. If the second series of statements
is given in the form of questions about the final state of the
world together with their correct answers, the agent should be able to learn from them and its
performance can be measured by the accuracy of its
answers. 

Even with this weak form of supervision, the system may learn basic
dynamical constraints about the world.
For example, it may learn that a person or object cannot be in two locations at the same time, or may learn
simple update rules such as incrementing and decrementing the
number of persons or objects in a room. 
It may also learn basic rules of approximate (logical) inference, such as the fact that objects belonging to the same category tend to have similar properties (light objects can be carried over from rooms to rooms for instance).

We propose to handle this scenario with a new kind of
memory-augmented neural network that uses a distributed memory and processor
architecture: the \modelname~(\modelabbrev). The model consists
of a fixed number of dynamic memory cells, each containing a vector
key $w_j$ and a vector value (or content) $h_j$. Each cell is
associated with its own ``processor'', a simple gated recurrent
network that may update the cell value given an input. If each
cell learns to represent a concept or entity in the world, one can imagine a gating
mechanism that, based on the key and content of the memory cells, will
only modify the cells that concern the entities mentioned in the
input. In the current version of the model, there is no direct
interaction between the memory cells, hence the system can be seen as
multiple identical processors functioning in parallel, with
distributed local memory. Alternatively, the \modelabbrev~can be seen as a
bank of gated RNNs (all sharing the same parameters), whose hidden
states correspond to latent concepts and attributes, and whose parameters describe the laws of the world according to which the attributes of objects are updated.
The sharing of these parameters reflects an invariance of these laws across object instances, similarly to how the weight tying scheme in a CNN reflects an invariance of image statistics across locations.
 Their hidden state
is updated only when new information relevant to their concept is
received, and remains otherwise unchanged. The keys used in the
addressing/gating mechanism also correspond to concepts or entities,
but are modified only during learning, not during inference.

The \modelabbrev~is able to solve all 20 bAbI question-answering tasks~\citep{babi}, a popular benchmark of story understanding, which to our knowledge sets a new state-of-the-art.
Our experiments also indicate that the model indeed maintains an internal representation of the simplified world in which the stories take place, and that the model does not limit itself to storing the aspects of the world required to answer a specific question.
We also introduce a new reasoning task which, unlike the bAbI tasks, requires a model to use a large number of supporting facts to answer the question, and show that the \modelabbrev~outperforms both LSTMs and Memory Networks~\citep{MemN2N} by a significant margin. It is also able to generalize to sequences longer than those seen during training.
Finally, our model also obtains competitive results on the
Children’s Book Test~\citep{CBT}, and performs best among models that read the text in a single pass before receiving
knowledge of the question.

\section{Model}

Our model is designed to process data in sequential form, and consists of three main parts: an input encoder, a dynamic memory and an output layer, which we now describe in detail.
We developed it in the context of question answering on short stories where the inputs are word sequences, but the model could be adapted to many other contexts.

\subsection{Input Encoder} The encoding layer summarizes an element of the input sequence with a vector of fixed length. 
Typically the input element at time $t$
 is a sequence of words, e.g. a sentence or window of words. 
One is free to choose the encoding module to be any standard sequence encoder, which is an active
area of research. Typical choices include
a bag-of-words (BoW) representation or the final state of a recurrent neural net (RNN)
 run over the sequence.
In this work, we use a simple encoder consisting of a learned multiplicative mask followed
by a summation. 
More precisely, let the input at time $t$ be a sequence of words with embeddings $\{e_1, ..., e_k\}$.
The vector representation of this input is then:

\begin{equation}
s_t = \sum_i f_i \odot e_i
\end{equation}

\begin{figure}[ht]
\begin{center}
\includegraphics[width=\textwidth]{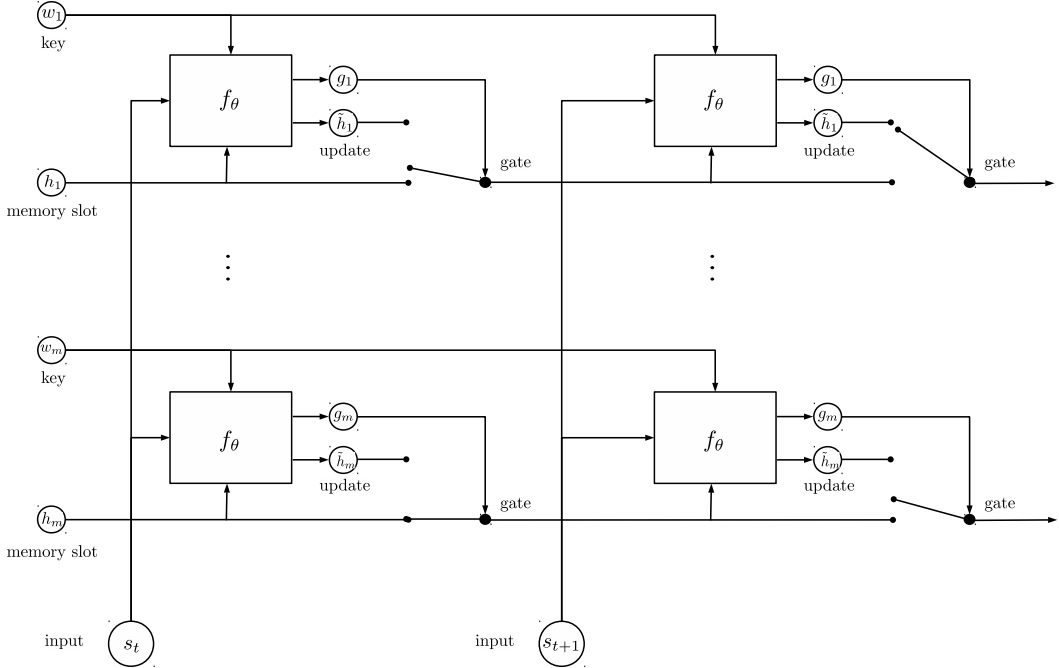}
\end{center}
\caption{Diagram of the \modelname's dynamic memory. Update equations 1 and 2 are represented by the module $f_\theta$, where $\theta$ is the set of trainable parameters. Equations 3 and 4 are represented by the gate, since they fullfill a similar function.}
\end{figure}

The same set of vectors $\{ f_1, ..., f_k \}$ are used at each time step and are learned jointly with the other parameters of the model.
Note that the model can choose to adopt a standard BoW representation by setting all weights in the multiplicative mask to 1, or can choose a positional encoding model as used in \citep{MemN2N}.

\subsection{Dynamic Memory}
The dynamic memory is a gated recurrent network with a (partially) block structured weight tying scheme. 
We divide the hidden states of the network into blocks $h_1,..., h_m$; the full hidden state is the concatenation of the $h_j$.  In the experiments below, $m$ is of the order of $5$ to $20$, and each block $h_j$ is of the order of $20$ to $100$ units.

At each time step $t$, the content of the hidden states $\{h_j\}$ (which we will call the $j$th memory) are updated using a set of key vectors $\{w_j\}$ and the encoded input $s_t$.
In its most general form, the update equations of our model are given by:

\begin{align}
\label{eq:less_simple_model}
g_j &\leftarrow \sigma(s_t^Th_j + s_t^T w_j) \\
\tilde{h_j} &\leftarrow \phi(Uh_j + Vw_j + Ws_t) \\
h_j &\leftarrow h_j + g_j \odot \tilde{h_j} \\
h_j &\leftarrow \frac{h_j}{||h_j||} 
\end{align}


Here $\sigma$ represents a sigmoid, $g_j$ is a gating function which determines how much the $j^{th}$ memory should be updated,
and $\tilde{h_j}$ is the new candidate value of the memory to be combined with the existing memory $h_j$.
The function $\phi$ can be chosen from any number of activation functions, in our experiments we use either parametric ReLU non-linearities \citep{PReLU} or the identity.
The matrices $U, V, W$ are typically trainable parameters of the model, and are shared between all the blocks.
They can also be fixed to certain values, such as the identity or zero, to yield a simpler model which we use in some of our experiments.

The gating function $g_j$ contains two terms: a ``content'' term $s_t^Th_j$ which causes the gate to open for memory slots whose content matches the input, and a ``location'' term $s_t^Tw_j$ which causes the gate to open for memory slots whose key matches the input.
The final normalization step allows the model to forget previous information. 
To see this, note that since the memories lie on the unit sphere, all information is contained in their phase. 
Adding any vector to a given memory (other than the memory itself) will decrease the cosine distance between the original memory and the updated one. 
Therefore, as new information is added, old information is forgotten.

\subsection{Output Module} 

Whenever the model is required to produce an output, it is presented with a query vector $q$. Specifically, the output is computed using the following equations:

\begin{equation}
\label{eq:output_less_simple}
\begin{split}
p_j &= \mbox{Softmax}(q^Th_j) \\
u &= \sum_j p_j h_j \\
y &= R\phi(q+Hu)
\end{split}
\end{equation}
The matrices $H$ and $R$ are additional trainable parameters of the model. The output module can be viewed as a one-hop Memory Network~\citep{MemN2N} with an additional non-linearity $\phi$ between the internal state and the decoder matrix.
If the memory slots correspond to specific words (as we will describe in the following section) which contain the answer, $p$ can be viewed as a distribution over potential answers and can be used to make a prediction directly or fed into a loss function, removing the need for the last two steps.

The entire model (all three components described above) is trained via backpropagation through time, receiving gradients from any time steps where the reader is required to produce an output, which are then propagated through the unrolled network.

\section{Motivating Example of Operation}

We now describe a motivating example of how our model
can perform reasoning on-the-fly as it is ingesting input sequences.
%
Let us suppose our model is reading a story, so the inputs are natural language sentences,
 and then it is required to answer questions about the story it has just read.

Our model is free to learn the key vectors $w_j$ for each memory $j$.
One choice the model could make is to associate a single memory (via the key) with each entity
in the story.
The memory slot corresponding to a person could encode that person's location, the objects they are carrying, or the people they are with, depending on what information is relevant for the task at hand. 
As new information is received indicating that objects are acquired or discarded,
 or the person changes location, their memory slot will change accordingly. Similarly useful updates
can be made for memories corresponding to object and location entities as well.

In fact, we could encode this choice of memories directly 
into our model, which we consider as a type of prior
knowledge.  
By tying the weights of the key vectors with the embeddings of specific words, we can encourage the model to record information about certain words occuring in the text which we believe to be important.
For example, given a list of named entities (which could be produced by a standard 
tagger), we could make the model have a separate memory slot for each entity.
We consider this ``tied'' variant in our experiments. 
Since the list of entities is independent of the training data, this variant can handle entities not seen in the training set, as long as their embeddings can be initialized in a reasonable way (such as pre-training on a larger corpus).

Now, consider that the model reads the following two sentences,
and the desired 
behavior of the gating function and update function at each memory as they are seen:

\begin{itemize}
\item \texttt{Mary picked up the ball.}
\item \texttt{Mary went to the garden.}
\end{itemize}

As the first sentence $s_t$ is ingested, and assuming memories encode entities,
we would like the gates of the memories corresponding to both ``Mary'' and ``ball'' 
to activate. This is possible  due to the location addressing term $s_t^Tw_j$ which uses the key
$w_j$. We expect that a well trained model would learn to do this.
The model would hence modify both the entry corresponding to ``Mary'' to indicate that 
she is now carrying the ball, and also the entry corresponding to ``ball'', to indicate
that it is being carried by Mary. 
When the second sentence is seen, we would like the model to again modify the ``Mary''
entry to indicate that she is now in the garden,
and also modify the ``ball'' entry to reflect its new location as well. 
Assuming the information for ``Mary'' is contained in the ``ball'' memory as described before,
the gate corresponding to ``ball'' can activate due to the content addressing term $s_t^Th_j$,
even though the word ``ball'' does not occur in the second sentence. 
As before, the gate corresponding to the ``Mary'' entry can open due to the second term.

If the gating function and update function have weights such that the steps above are executed,
then the memory will be in a state where questions such as
``Where is the ball?'' or ``Where is Mary?'' can be answered from the values of relevant memories,
without the need for further complex reasoning.

\section{Related Work} \label{sec:related}

The \modelabbrev~is related to gated recurrent models such as the LSTM \citep{LSTM} and GRU \citep{GRU}, which also use gates to fix or modify the information stored in the hidden state. However, these models use scalar memory cells with full interactions between them, whereas ours has separate memory slots which could be seen as groups of hidden units with tied weights in the gating and update functions.
Another important difference is the content-based matching term between the input and hidden state, which is not present in these models. 

Our model also shares some similarities with the DNC/NTM framework of \citep{NTM, graves2016hybrid}. There, as in our model, a block of hidden states acts as a set of read-writeable memories. On the other hand, the DNC has a relatively sophisticated controller network (such as an LSTM) which reads an input and outputs a number of interface vectors (such as keys and weightings) which are then combined via a softmax to read from and write to the external memory matrix. 
In contrast, our model can be viewed as a set of separate recurrent models whose hidden states store the memory slots. These hidden states are either fixed by the gates, or modified through a simple RNN-style update. The bulk of the reasoning is thus performed by these parallel recurrent models, rather than through a central controller.  Moreover, instead of using a softmax, our model uses an independent gate for writing to each memory.  


Our model is similar to a Memory Network and its variants \citep{MemNet, MemN2N, chandar2016hierarchical, miller2016key} in the way it produces an output using a softmax over blocks of hidden states, and our encoding layer is inspired by techniques used in those works. However, Memory Networks explicitly
store the entire input sequence in memory, and then sequentially update a controller's hidden state via a softmax gating over the memories. In contrast, our model keeps a fixed number of blocks of hiddens as memories and updates each block with an independent gated RNN. 
The Dynamic Memory Network of \citep{DMN} also performs updates via a recurrent model, however it links memories to input tokens and updates them sequentially rather than in parallel.


The weight tying scheme and the parallel gated RNNs recall the gated graph network of \citep{GGSNN}.  
If we interpret our work in that context, the ``graph'' is just a set of vertices with no edges; our gating mechanism is also somewhat different than the one they use.   
The CommNN model of \citep{CommNN}, the Interaction Network of \citep{InteractionNet}, the Neural Physics Engine of \citep{Chang2016} and the model of \citep{FragkiadakiALM15} also use a set of parallel recurrent models with tied weights, but differ from our model in their use of inter-network communication and the lack of a gating mechanism.

Finally, there is another class of recent models that have a writeable memory
arranged as (unbounded) stacks, linked lists or queues \citep{joulin2015inferring, grefenstette2015learning}.  Our model is different from these in that we use a key-value pair array instead of a stack, and in the experiments in this work, the array is of fixed size.

\section{Experiments}

In this section we evaluate our model on three different datasets. 
Training details common to all experiments can be found in Appendix \ref{sec:training_details}.

\subsection{Synthetic World Model Task}

We first study our model's properties on a toy task designed to measure the ability to keep a world model in memory.
In this task two agents are initially placed randomly on an $10 \times 10$ grid, and at each time step a randomly chosen agent either changes direction or moves ahead.
After a certain number of time steps, the model is required to provide the locations of each of the agents, thus revealing its internal world model (details can be found in Appendix~\ref{sec:world_model}). This task is challenging because the model must combine up to $T-2$ supporting facts in order to answer the question correctly, and must also keep the locations of both agents in memory and update them at different times. 

\begin{table}
\subfloat[]{
\scalebox{0.95}{
\begin{tabular}{llll}
  \hline
  Model & $T=10$ & $T=20$ & $T=40$ \\
  \hline
  MemN2N & 0.09 & 0.633 & 0.896 \\
  LSTM & 0 & 0.157 & 0.226 \\
  \modelabbrev & 0 & 0 & 0 \\
  \hline 
\end{tabular}
}}
\hspace{1.5mm}
\subfloat[]{
\scalebox{0.95}{
\begin{tabular}{llllllll}
  \hline 
  T & $20$ & $30$ & $40$ & $50$ & $60$ & $70$ & $80$ \\
  \hline 
  Error & 0 & 0 & 0 & 0.01 & 0.03 & 0.05 & 0.08 \\
  \\ \\ \hline 
\end{tabular}
}}
\caption{a) Error of different models on the World Model Task. b) Generalization of an EntNet trained up to $T=20$. All errors range from 0 to 1.}
\end{table}

We compared the performance of a MemN2N, LSTM and \modelabbrev. 
For the MemN2N, we set the number of hops equal to $T-2$ and the embedding dimension to $d=20$.
The \modelabbrev~ had embedding dimension $d=20$ and 5 memory slots, and the LSTM had $50$ hidden units which resulted in it having significantly more parameters than the other two models.
For each model, we repeated the experiment with 5 different initializations and reported the best performance. 
All models were trained with ADAM \citep{ADAM} with initial learning rates set by grid search over $\{0.1, 0.01, 0.001 \}$ and divided by 2 every 10,000 updates.
Table 1a shows the results. The MemN2N has the worst performance, which degrades quickly as the length of the sequence increases. 
The LSTM performs better, but still loses accuracy as the length of the sequence increases. 
In contrast, the \modelabbrev~ is able to solve the task in all cases. 

The ability to generalize to sequences longer than those seen during training is a desirable property, which suggests that the network has learned the dynamics of the world it is trying to model. It also means the model can be trained less expensively. 
To study this, we trained an \modelabbrev~on variable length sequences between 1 and 20, and evaluated it on different length sequences longer than 20. 
Results are shown in Table 1b. We see that the model is able to achieve good performance several times past its training horizon.

\subsection{bAbI Tasks}

We next evaluate our model on the bAbI tasks, which are a collection of 20 synthetic question-answering datasets first introduced in \citep{babi} designed to test a wide variety of reasoning abilities. 
They have since become a benchmark for memory-augmented neural networks and most of the related methods
 described in Section \ref{sec:related} have been tested on them.
Performance is measured using two metrics: the average error across all tasks, and the number of failed tasks (more than $5\%$ error).
We used version 1.2 of the dataset with 10k samples. 
\footnote{Code to reproduce these experiments can be found at \\ \texttt{https://github.com/facebook/MemNN/tree/master/EntNet-babi}.}

\textbf{Training Details} We used a similar training setup as \citep{MemN2N}. All models were trained with ADAM using a learning rate of $\eta=0.01$, which was divided by 2 every 25 epochs until 200 epochs were reached. 
Copying previous works \citep{MemN2N,DMN}, the capacity of the memory was limited to the most recent 70 sentences, except for task 3 which was limited to 130 sentences. 
Due to the high variance in model performance for some tasks, for each task we conducted 10 runs with different initializations and picked the best model based on performance on the validation set, as it has been done in previous work. 
In all experiments, our model had embedding dimension size $d=100$ and 20 memory slots. 

\begin{table}[t!]
\caption{Results on bAbI Tasks with 10k training samples.}
\label{babi-results}
\begin{center}
\resizebox{0.9\textwidth}{!}{
\begin{tabular}{lllllll}
\multicolumn{1}{c}{\bf }  &\multicolumn{2}{c}{}
\\ \hline \\
Task & NTM & D-NTM & MemN2N & DNC & DMN+ & \modelabbrev \\
\\ \hline \\
1: 1 supporting fact       & 31.5  & 4.4 & 0 & 0 & 0 & 0 \\
2: 2 supporting facts      & 54.5  & 27.5 & 0.3 & 0.4 & 0.3 & 0.1 \\ 
3: 3 supporting facts      & 43.9  & 71.3 & 2.1 & 1.8 & 1.1 & 4.1 \\
4: 2 argument relations    & 0     & 0 & 0 & 0 & 0 & 0 \\
5: 3 argument relations    & 0.8   & 1.7 & 0.8 & 0.8 & 0.5 & 0.3 \\
6: yes/no questions        & 17.1  & 1.5 & 0.1 & 0 & 0 & 0.2 \\
7: counting                & 17.8  & 6.0 & 2.0 & 0.6 & 2.4 & 0\\
8: lists/sets              & 13.8  & 1.7 & 0.9 & 0.3 & 0.0 & 0.5\\
9: simple negation         & 16.4  & 0.6 & 0.3 & 0.2 & 0.0 & 0.1 \\
10: indefinite knowledge   & 16.6  & 19.8 & 0 & 0.2 & 0 & 0.6 \\
11: basic coreference      & 15.2  & 0 & 0.0 & 0 & 0.0 & 0.3\\
12: conjunction            & 8.9   & 6.2 & 0 & 0 & 0.2 & 0 \\
13: compound coreference   & 7.4   & 7.5 & 0 & 0 & 0 & 1.3 \\
14: time reasoning         & 24.2  & 17.5 & 0.2 & 0.4 & 0.2 & 0 \\
15: basic deduction        & 47.0  & 0 & 0 & 0 & 0 & 0 \\
16: basic induction        & 53.6  & 49.6 & 51.8 & 55.1 & 45.3 & 0.2\\
17: positional reasoning   & 25.5  & 1.2  & 18.6 & 12.0 & 4.2 & 0.5 \\
18: size reasoning         & 2.2   & 0.2 & 5.3 & 0.8 & 2.1 & 0.3 \\
19: path finding           & 4.3   & 39.5 & 2.3 & 3.9 & 0.0 & 2.3\\
20: agent's motivation     & 1.5   & 0 & 0 & 0 & 0 & 0\\
\\ \hline \\
Failed Tasks ($>5\%$ error): & 16 & 9 & 3 & 2 & 1 & \textbf{0} \\
Mean Error: & 20.1 & 12.8 & 4.2 & 3.8 & 2.8 & \textbf{0.5} \\
\end{tabular}}
\end{center}
\end{table}

In Table \ref{babi-results} we compare our model to various other state-of-the-art models in the literature: the larger MemN2N reported in the appendix of \citep{MemN2N}, the Dynamic Memory Network of \citep{DMN}, the Dynamic Neural Turing Machine \citep{DNTM}, the Neural Turing Machine \citep{NTM} and the Differentiable Neural Computer \citep{graves2016hybrid}.
Our model is able to solve all the tasks, outperforming the other models in terms of both the number of solved tasks and the average error. 

To analyze what kind of representations our model can learn, we conducted an additional experiment on Task 2 using a simple BoW sentence encoding and key vectors which were tied to entity embeddings. 
This was designed to make the model more interpretable, since the weight tying forces memory slots to encode information about specific entities. \footnote{For most tasks including this one, tying key vectors did not significantly change performance, although it hurt in a few cases (see Appendix~\ref{sec:app_babi}). Therefore we did not apply it in Table \ref{babi-results}}
After training, we ran the model over a story and computed the cosine distance between $\phi(Hh_j)$ and each row $r_i$ of the decoder matrix $R$. This gave us a score which measures the affinity between a given memory slot and each word in the vocabulary. 
Table \ref{world-model-viz} shows the nearest neighboring words for each memory slot (which itself corresponds to an entity). 
We see that the model has indeed stored locations of all of the objects and characters in its memory slots which reflect the final state of the story. In particular, it has the correct answer readily stored in the memory slot of the entity being inquired about (the milk). It also has correct location information about all other non-location entities stored in the appropriate memory slots. 
Note that it does not store useful or correct information in the memory slots corresponding to locations, most likely because this task does not contain questions about locations (such as ``who is in the kitchen?'').

\begin{table}[t]
  \caption{On the left, the network's final ``world model'' after reading the story on the right. First and second nearest neighbors from each memory slot are shown, along with their cosine distance.}
  \label{world-model-viz}
  \resizebox{0.99\textwidth}{!}{
    \begin{minipage}{0.5\textwidth}
      \centering
      \resizebox{0.9\textwidth}{!}{
        \begin{tabular}{lllllll}
          \\ Key & 1-NN & 2-NN \\
          \hline \\
          football & hallway (0.135) & dropped (0.056) \\ 
          milk & garden (0.111) & took (0.011) \\
          john & kitchen (0.501) & dropped (0.027) \\
          mary & garden (0.442) & took (0.034) \\
          sandra & hallway (0.394) & kitchen (0.121) \\
          daniel & hallway (0.689) & to (0.076) \\
          bedroom & hallway (0.367) & dropped (0.075) \\
          kitchen & kitchen (0.483) & daniel (0.029) \\
          garden & garden (0.281) & where (0.026) \\
          hallway & hallway (0.475) & left (0.060) \\
          \\
          \\ 
          \\
          \\
          \hline
        \end{tabular}
      }
    \end{minipage}
    \begin{minipage}{0.5\textwidth}
      \centering
      \resizebox{0.9\textwidth}{!}{
        \begin{tabular}{lllllll}
          \\ Story \\
          \hline \\
          \texttt{mary got the milk there} \\ 
          \texttt{john moved to the bedroom} \\
          \texttt{sandra went back to the kitchen} \\
          \texttt{mary travelled to the hallway} \\
          \texttt{john got the football there} \\
          \texttt{john went to the hallway}\\
          \texttt{john put down the football} \\
          \texttt{mary went to the garden} \\
          \texttt{john went to the kitchen} \\
          \texttt{sandra travelled to the hallway} \\
          \texttt{daniel went to the hallway} \\
          \texttt{mary discarded the milk}\\
          \texttt{where is the milk ?}\\
          \texttt{answer: garden} \\
          \hline
        \end{tabular}
      }
    \end{minipage}
  }
\end{table}

\subsection{Children's Book Test (CBT)}

We next evaluated our model on the Children's Book Test \citep{CBT}, which is a semantic language modeling (sentence completion) 
benchmark built from children's books that are freely available from Project Gutenberg \footnote{www.gutenberg.org}. 
Models are required to read 20 consecutive sentences from a given story and use this context to fill in a missing word from the 21st sentence. 
More specifically, each sample consists of a tuple $(S,q,C,a)$ where $S$ is the story consisting of 20 sentences, $Q$ is the 21st sentence with one word replaced by a special blank token, $C$ is a set of 10 candidate answers of the same type as the missing word (for example, common nouns or named entities), and $a$ is the true answer (which is always contained in $C$).

It was shown in \citep{CBT} that methods with limited memory such as LSTMs perform well on more frequent, syntax based
words such as prepositions and verbs, being similar to human performance, but poorly relative to humans on more semantically meaningful words such as named entities and common nouns.
Therefore, most recent methods have been evaluated on the Named Entity and Common Noun subtasks, since they better test the ability of a model to make use of wider contextual information. 

\textbf{Training Details} 
We adopted the same window memory approach used in \citep{CBT}, where each input corresponds to a window of text from $\{w_{(i-b-1/2)}...w_i...w_{(i+(b-1)/2)}\}$ centered at a candidate $w_i \in C$. In our experiments we set $b=5$. 
All models were trained using standard stochastic gradient descent (SGD) with a fixed learning rate of 0.001. 
We used separate input encodings for the update and gating functions, and applied a dropout rate of $0.5$ to the word embedding dimensions.
Key embeddings were tied to the embeddings of the candidate words, resulting in 10 hidden blocks, one per member of $C$.
Due to the weight tying, we did not need a decoder matrix and used the distribution over candidates to directly produce a prediction, as described in Section 3.

We found that a simpler version of the model worked best, with $U = V = 0$, $W = I$ and $\phi$ equal to the identity.
We also removed the normalization step in this simplified model, which we found to hurt performance.
This can be explained by the fact that the maximum frequency baseline model in \citep{CBT} has performance which is significantly higher than random, and including the normalization step hides this useful frequency-based information.

\begin{table}[t]
\caption{Accuracy on CBT test set. Single-pass models encode the document before seeing the query, multi-pass models have access to the query at read time.}
\label{cbt-results}
\begin{center}
\resizebox{\textwidth}{!}{
\begin{tabular}{lllll}
\hline \\
& Model & Named Entities & Common Nouns \\
\\ \hline \\
\multirow{4}{*}{Single Pass}
& Kneser-Ney Language Model + cache & 0.439 & 0.577 \\
& LSTMs (context + query) & 0.418 & 0.560 \\
& Window LSTM & 0.436 & 0.582 \\
& EntNet (general) & 0.484 & 0.540 \\
& EntNet (simple) & \textbf{0.616} & \textbf{0.588} \\
\\ \hline \\
\multirow{6}{*}{Multi Pass}
& MemNN & 0.493 & 0.554 \\
& MemNN + self-sup. & 0.666 & 0.630 \\
& Attention Sum Reader \citep{ASReader} & 0.686 & 0.634 \\ 
& Gated-Attention Reader \citep{GAReader} & 0.690 & 0.639 \\
& EpiReader \citep{EpiReader} & 0.697 & 0.674 \\
& AoA Reader \citep{AoAReader} & 0.720 & 0.694 \\
& NSE Adaptive Computation \citep{NSE} & \textbf{0.732} & \textbf{0.714} \\
\\ \hline 
\end{tabular}
}
\end{center}
\end{table}

\textbf{Results} 
We draw a distinction between two setups: the single-pass setup, where the model must read the story and query in order and immediately produce an output, and the multi-pass setup, where the model can use the query to perform attention over the story. 
The first setup is more challenging because the model does not know beforehand which query it will be presented with, and must learn to retain information which is useful for a wide variety of potential queries. 
For this reason it can be viewed as a test of the model's ability to construct a general-purpose representation of the current state of the story. 
The second setup leverages all available information, and allows the model to use knowledge of which question will be asked when it reads the story.

In Table \ref{cbt-results}, we show the performance of the general EntNet, the simplified EntNet, as well as other single-pass models taken from \citep{CBT}.
The general EntNet performs better than the LSTMs and $n$-gram model on the Named Entities Task, but lags behind on the Common Nouns task. 
The simplified EntNet outperforms all other single-pass models on both tasks, and also performs better than the Memory Network which does not use the self-supervision heuristic. 
However, there is still a performance gap when compared to more sophisticated machine comprehension models, many of which perform multiple layers of attention over the story using query knowledge.
The fact that the simplified \modelabbrev~is able to obtain decent performance is encouraging since it indicates that the model is able to build an internal representation of the story which it can then use to answer a relatively diverse set of queries.

\section{Conclusion}

Two closely related challenges in artificial intelligence are designing models which can maintain an estimate of the state of a world with complex dynamics over long timescales, and models which can predict the forward evolution of the state of the world from partial observation.
In this paper, we introduced the \modelname, a new model that makes a
promising step towards the first goal. Our model is able to accurately track the world state while reading text
stories, which enables it to set a new
state-of-the-art on the bAbI tasks, the competitive benchmark of story
understanding, by being the first model to solve them all.
We also showed that our model is able to capture simple dynamics over long timescales, and is able to perform competitively on a real-world dataset.

Although our model was able to solve all the bAbI tasks using 10k training samples, we found that performance dropped considerably when using only 1k samples (see Appendix). 
Most recent work on the bAbI tasks has focused on the 10k samples setting, and we would like to emphasize that solving them in the 1k samples setting remains an open problem which will require improving the sample efficiency of reasoning models, including ours.

Recent works have made some progress towards the second goal of forward modeling, for instance in capturing simple physics ~\citep{lerer-icml-2016}, predicting future frames in video ~\citep{mathieu-iclr-2016} or responses in dialog ~\citep{weston16}. 
Although we have only applied our model to tasks with textual inputs in this work, the architecture is general and future work should investigate how to combine the \modelabbrev's tracking abilities with such predictive models.


%

\bibliography{iclr2016_conference}
\bibliographystyle{iclr2016_conference}

\appendix

\section{Training Details}\label{sec:training_details}

All models were implemented using Torch \citep{Torch7}.
In all experiments, we initialized our model by drawing weights from a Gaussian distribution with mean zero and standard deviation 0.1, except for the PReLU slopes and encoder weights which were initialized to 1. 
Note that the PReLU initialization is related to two of the heuristics used in \citep{MemN2N}, namely starting training with a purely linear model, and adding non-linearities to half of the hidden units. Our initialization allows the model to choose when and how much to enter the non-linear regime.
Initializing the encoder weights to 1 corresponds to beginning with a BoW encoding, which the model can then choose to modify.
The initial values of the memory slots were initialized to the key values, which we found to help performance.
Optimization was done with SGD or ADAM using minibatches of size 32, and gradients with norm greater than 40 were clipped to 40. 
A null symbol whose embedding was constrained to be zero was used to pad all sentences or windows to a fixed size.

\section{Details of World Model Experiments}\label{sec:world_model}

Two agents are initially placed at random on a $10 \times 10$ grid with 100 distinct locations $\{(1,1),(1,2),...(9,10),(10,10) \}$.
At each time step an agent is chosen at random. 
There are two types of actions: the agent can face a given direction, or can move a number of steps ahead.
Actions are sampled until a legal action is found by either choosing to change direction or move with equal probability. If they change direction, the direction is chosen between north, south, east and west with equal probability. If they move, the number of steps is randomly chosen between 1 and 5. A legal action is one which does not place the agent off the grid.
Stories are given to the network in textual form, an example of which is below. The first action after each agent is placed on the grid is to face a given direction. 
Therefore, the maximum number of actions made by one agent is $T-2$. 
The network learns word embeddings for all words in the vocabulary such as locations, agent identifiers and actions.
At question time, the model must predict the correct answer (which will always be a location) from all the tokens in the vocabulary.

\texttt{agent1 is at (2,8) \\
 agent1 faces-N \\
 agent2 is at (9,7) \\
 agent2 faces-N \\
 agent2 moves-2 \\
 agent2 faces-E \\
 agent2 moves-1 \\
 agent1 moves-1  \\
 agent2 faces-S   \\
 agent2 moves-5   \\
 Q1: where is agent1 ? \\
 Q2: where is agent2 ? \\
 A1: (2,9) \\
 A2: (10,4) \\
}

\section{Additional Results on bAbI Tasks}\label{sec:app_babi}

We provide some additional experiments on the bAbI tasks, in order to better understand the influence of architecture, weight tying, and amount of training data. Table \ref{bow-babi} shows results when a simple BoW encoding is used for the inputs. Here, the \modelabbrev~still performs better than a MemN2N which uses the same encoding scheme, indicating that the architecture has an important effect.
Tying the key vectors to entities did not help, and hurt performance for some tasks. Table \ref{1k-samples} shows results when using only 1k training samples. In this setting, the \modelabbrev~ performs worse than the MemN2N.

Table \ref{joint} shows results for the EntNet and the DNC when models are trained on all tasks jointly. 
We report results for the mean performance across different random seeds (20 for the DNC, 5 for the EntNet), as well as the performance for the single best seed (measured by validation error).
The DNC results for mean performance were taken from the appendix of \cite{graves2016hybrid}.
The DNC has better performance in terms of the best seed, but also exhibits high variation across seeds, indicating that many different runs are required to achieve good performance.
The EntNet exhibits less variation across runs and is able to solve more tasks consistently.
Note that Table \ref{babi-results} reports DNC results with joint training, since results when training on each task separately were not available.

\begin{table}[htp!]
\caption{Error rates on bAbI Tasks with inputs are encoded using BoW. ``Tied'' refers to the case where key vectors are tied with entity embeddings.}
\label{bow-babi}
\begin{center}
\begin{tabular}{llllll}
\multicolumn{1}{c}{\bf }  &\multicolumn{2}{c}{}
\\ \hline \\
Task & MemN2N & \modelabbrev-tied & \modelabbrev\\
\\ \hline \\
1: 1 supporting fact         & 0 & 0 & 0\\
2: 2 supporting facts        & 0.6 & 3.0 & 1.2\\ 
3: 3 supporting facts        & 7 & 9.6 & 9.0 \\
4: 2 argument relations      & 32.6 & 33.8 & 31.8 \\
5: 3 argument relations      & 10.2 & 1.7 & 3.5\\
6: yes/no questions          & 0.2 & 0 & 0 \\
7: counting & 10.6 & 0.5 & 0.5\\
8: lists/sets & 2.6 & 0.1 & 0.3\\
9: simple negation & 0.3 & 0 & 0 \\
10: indefinite knowledge & 0.5 & 0 & 0 \\
11: basic coreference & 0 & 0.3 & 0\\
12: conjunction & 0 & 0 & 0\\
13: compound coreference & 0 & 0.2 & 0.4 \\
14: time reasoning & 0.1 & 6.2 & 0.1 \\
15: basic deduction & 11.4 & 12.5 & 12.1 \\
16: basic induction & 52.9 & 46.5 & 0\\
17: positional reasoning & 39.3 & 40.5 & 40.5 \\
18: size reasoning & 40.5 & 44.2 & 45.7 \\
19: path finding & 74.4 & 75.1 & 74.0\\
20: agent's motivation & 0 & 0 & 0\\
\\ \hline \\
Failed Tasks ($> 5\%$): & 9 & 8 & \textbf{6}\\
Mean Error: & 15.6 & 13.7 & \textbf{10.9}\\
\end{tabular}
\end{center}
\end{table}

\begin{table}[t]
\caption{Results on bAbI Tasks with 1k samples.}
\label{1k-samples}
\begin{center}
\begin{tabular}{llllll}
\multicolumn{1}{c}{\bf }  &\multicolumn{2}{c}{}
\\ \hline \\
Task & MemN2N & \modelabbrev \\
\\ \hline \\
1: 1 supporting fact & 0 & 0.7        \\
2: 2 supporting facts & 8.3 & 56.4       \\
3: 3 supporting facts & 40.3 & 69.7       \\
4: 2 argument relations & 2.8 & 1.4      \\
5: 3 argument relations & 13.1 & 4.6      \\
6: yes/no questions & 7.6 & 30.0         \\
7: counting & 17.3 & 22.3 \\
8: lists/sets & 10.0 & 19.2 \\
9: simple negation & 13.2 & 31.5 \\
10: indefinite knowledge & 15.1 & 15.6 \\
11: basic coreference & 0.9 & 8.0 \\
12: conjunction & 0.2 & 0.8\\
13: compound coreference & 0.4 & 9.0 \\
14: time reasoning & 1.7 & 62.9\\
15: basic deduction & 0 & 57.8\\
16: basic induction & 1.3 & 53.2 \\
17: positional reasoning & 51.0 & 46.4 \\
18: size reasoning & 11.1 & 8.8\\
19: path finding  & 82.8 & 90.4 \\
20: agent's motivation & 0 & 2.6\\
\\ \hline \\
Failed Tasks ($> 5\%$): & \textbf{11} & 15\\
Mean Error: & \textbf{13.9} & 29.6\\
\end{tabular}
\end{center}
\end{table}

\begin{table}[t]
\caption{Results on bAbI Tasks with 10k samples and joint training on all tasks.}
\label{joint}
\begin{center}
\begin{tabular}{lll|lll}
\hline \\
\multicolumn{1}{c}{} & \multicolumn{2}{c}{All Seeds} & \multicolumn{2}{c}{Best Seed} \\
\hline 
Task & DNC & \modelabbrev & DNC &  \modelabbrev \\
\hline 
1: 1 supporting fact & $9.0 \pm 12.6$ & $0 \pm 0.1$ & 0 & 0.1       \\
2: 2 supporting facts & $39.2 \pm 20.5$ & $15.3 \pm 15.7$ & 0.4 & 2.8       \\
3: 3 supporting facts & $39.6 \pm 16.4$ & $29.3 \pm 26.3 $ & 1.8 & 10.6        \\
4: 2 argument relations & $0.4 \pm 0.7$ & $0.1 \pm 0.1$ & 0 & 0      \\
5: 3 argument relations & $1.5 \pm 1.0$ & $0.4 \pm 0.3$ & 0.8 & 0.4     \\
6: yes/no questions & $6.9 \pm 7.5$ & $0.6 \pm 0.8$ & 0 & 0.3         \\
7: counting & $9.8 \pm 7.0$ & $1.8 \pm 1.1$ & 0.6 & 0.8 \\
8: lists/sets & $5.5 \pm 5.9$ & $1.5 \pm 1.2$ & 0.3 & 0.1 \\
9: simple negation & $7.7 \pm 8.3$ & $0 \pm 0.1$ & 0.2 & 0 \\
10: indefinite knowledge & $9.6 \pm 11.4$ & $0.1 \pm 0.2$ & 0.2 & 0 \\
11: basic coreference & $3.3 \pm 5.7$ & $0.2 \pm 0.2$ & 0 & 0\\
12: conjunction & $5.0 \pm 6.3$ & $0 \pm 0$ & 0 & 0\\
13: compound coreference & $3.1 \pm 3.6$ & $0 \pm 0.1$ & 0 & 0 \\
14: time reasoning & $11.0 \pm 7.5$ & $7.3 \pm 4.5$ & 0.4 & 3.6 \\
15: basic deduction & $27.2 \pm 20.1$ & $3.6 \pm 8.1$ & 0 & 0 \\
16: basic induction & $53.6 \pm 1.9$ & $53.3 \pm 1.2 $ & 55.1 & 52.1 \\
17: positional reasoning & $32.4 \pm 8.0$ & $8.8 \pm 3.8$ & 12.0 & 11.7 \\
18: size reasoning & $4.2 \pm 1.8$ & $1.3 \pm 0.9$ & 0.8 & 2.1 \\
19: path finding  & $64.6 \pm 37.4$ & $70.4 \pm 6.1$ & 3.9 & 63.0 \\
20: agent's motivation & $0.0 \pm 0.1$ & $0 \pm 0$ & 0 & 0\\
\hline
Failed Tasks ($> 5\%$): & $11.2 \pm 5.4$  & $\mathbf{5 \pm 1.2}$ & \textbf{2} & 4 \\
Mean Error: & $16.7 \pm 7.6$ & $\mathbf{9.7 \pm 2.6}$ & \textbf{3.8} & 7.38 \\
\end{tabular}
\end{center}
\end{table}

\end{document}